\documentclass[11pt,a4paper]{article}
\usepackage[hyperref]{acl2020}
\usepackage{times}
\usepackage{latexsym}

\usepackage{microtype}

\usepackage{array} 
\usepackage{booktabs}  
\usepackage{threeparttable}  
\usepackage{multicol}  
\usepackage{multirow}  
\usepackage{natbib}
\usepackage{epsfig}
\usepackage{algorithm}
\usepackage{algorithmic}
\usepackage{amssymb}
\usepackage[utf8]{inputenc}
\usepackage{cleveref}
\usepackage{verbatim}
\crefname{section}{§}{§§}
\Crefname{section}{§}{§§}

\usepackage{bm}
\usepackage{subcaption}

\newcommand*{\affmark}[1][*]{\textsuperscript{#1}}
\newcommand*{\affaddr}[1]{#1} 

\aclfinalcopy 

\title{Dynamic Memory Induction Networks for Few-Shot Text Classification}

\author{
  Ruiying Geng\affmark[1], Binhua Li\affmark[1], Yongbin Li\affmark[1]\thanks{$^*$Corresponding author.} , Jian Sun\affmark[1], Xiaodan Zhu\affmark[2]\\
\affaddr{
 \affaddr{\affmark[1] Alibaba Group, Beijing}\\
   \affaddr{\affmark[2] Ingenuity Labs Research Institute \& ECE, Queen's University}\\
{\tt \{ruiying.gry,binhua.lbh,shuide.lyb,jian.sun\}@alibaba-inc.com}\\
    {\tt zhu2048@gmail.com    } \\
  }
}

\date{}

\begin{document}
\maketitle

\begin{abstract}
This paper proposes Dynamic Memory Induction Networks (DMIN) for few-shot text classification. The model utilizes dynamic routing to provide more flexibility to memory-based few-shot learning in order to better adapt the support sets, which is a critical capacity of few-shot classification models. Based on that, we further develop  induction models with query information, aiming to enhance the generalization ability of meta-learning. The proposed model achieves new state-of-the-art results on the miniRCV1 and ODIC dataset, improving the best performance (accuracy) by 2$\sim$4$\%$. Detailed analysis is further performed to show the effectiveness of each component.
\end{abstract}



\section{Introduction}
Few-shot text classification, which requires models to perform classification with a limited number of training instances, is important for many applications but yet remains to be a challenging task. 
Early studies on few-shot learning \citep{salamon2017deep} employ data augmentation and regularization techniques to alleviate overfitting caused by data sparseness. 
More recent research leverages meta-learning \citep{finn2017model, zhang2018metagan, sun2019meta} to extract transferable knowledge among meta-tasks in meta episodes. 

A key challenge for few-shot text classification is inducing class-level representation from support sets \citep{gao2019hybrid}, in which key information is often lost when switching between meta-tasks. Recent solutions \citep{gidaris2018dynamic} leverage a memory component to maintain models' learning experience, e.g., by finding from a supervised stage the content that is similar to the unseen classes, leading to the state-of-the-art performance. However, the memory weights are static during inference and the capability of the model is still limited when adapted to new classes. Another prominent challenge is the instance-level diversity caused by various reasons \citep{gao2019hybrid, geng-etal-2019-induction}, resulting in the difficulty of finding a fixed prototype for a class~\citep{pmlr-v97-allen19b}. Recent research has shown that models can benefit from query-aware methods \citep{gao2019hybrid}. 








In this paper we propose Dynamic Memory Induction Networks (DMIN) to further tackle the above challenges. DMIN utilizes dynamic routing \citep{sabour2017dynamic, geng-etal-2019-induction} to render more flexibility to memory-based few-shot learning \citep{gidaris2018dynamic} in order to better adapt the support sets, by leveraging the routing component's capacity in automatically adjusting the coupling coefficients during and after training. Based on that, we further develop induction models with query information to identify, among diverse instances in support sets, the sample vectors that are more relevant to the query. These two modules are jointly learned in DMIN.

The proposed model achieves new state-of-the-art results on the miniRCV1 and ODIC datasets, improving the best performance by 2$\sim$4$\%$ accuracy. We perform detailed analysis to further show how the proposed network achieves the improvement.





\section{Related Work}
\label{related:appendix}
Few-shot learning has been studied in early work such as \citep{fe2003bayesian,fei2006one} and more recent work \citep{ba2016using, santoro2016meta, munkhdalai2017meta, ravi2016optimization, mishra2017simple,finn2017model, 
vinyals2016matching, snell2017prototypical, sung2018learning, pmlr-v97-allen19b}. Researchers have also investigated few-shot learning in various NLP tasks \citep{dou2019investigating, wu2019learning, gu2018meta, chen-etal-2019-meta, obamuyide-vlachos-2019-model,hu-etal-2019-shot}, including text classification \citep{yu2018diverse, rios2018few, xu2019open, geng-etal-2019-induction, gao2019hybrid, ye-ling-2019-multi}.


Memory mechanism has shown to be very effective in many NLP tasks \citep{tang2016aspect, das-EtAl:2017:Short, madotto2018mem2seq}. In the few-shot learning scenario, researchers have applied memory networks to store the encoded contextual information in each meta episode \citep{santoro2016meta, cai2018memory, kaiser2017learning}. Specifically \citet{qi2018low} and \citet{gidaris2018dynamic} build a two-stage training procedure and regard the supervisely learned class representation as a memory component.

\section{Dynamic Memory Induction Network}
\subsection{Overall Architecture}
\begin{figure}[t]
\centering
\small
\includegraphics[width=7.5cm]{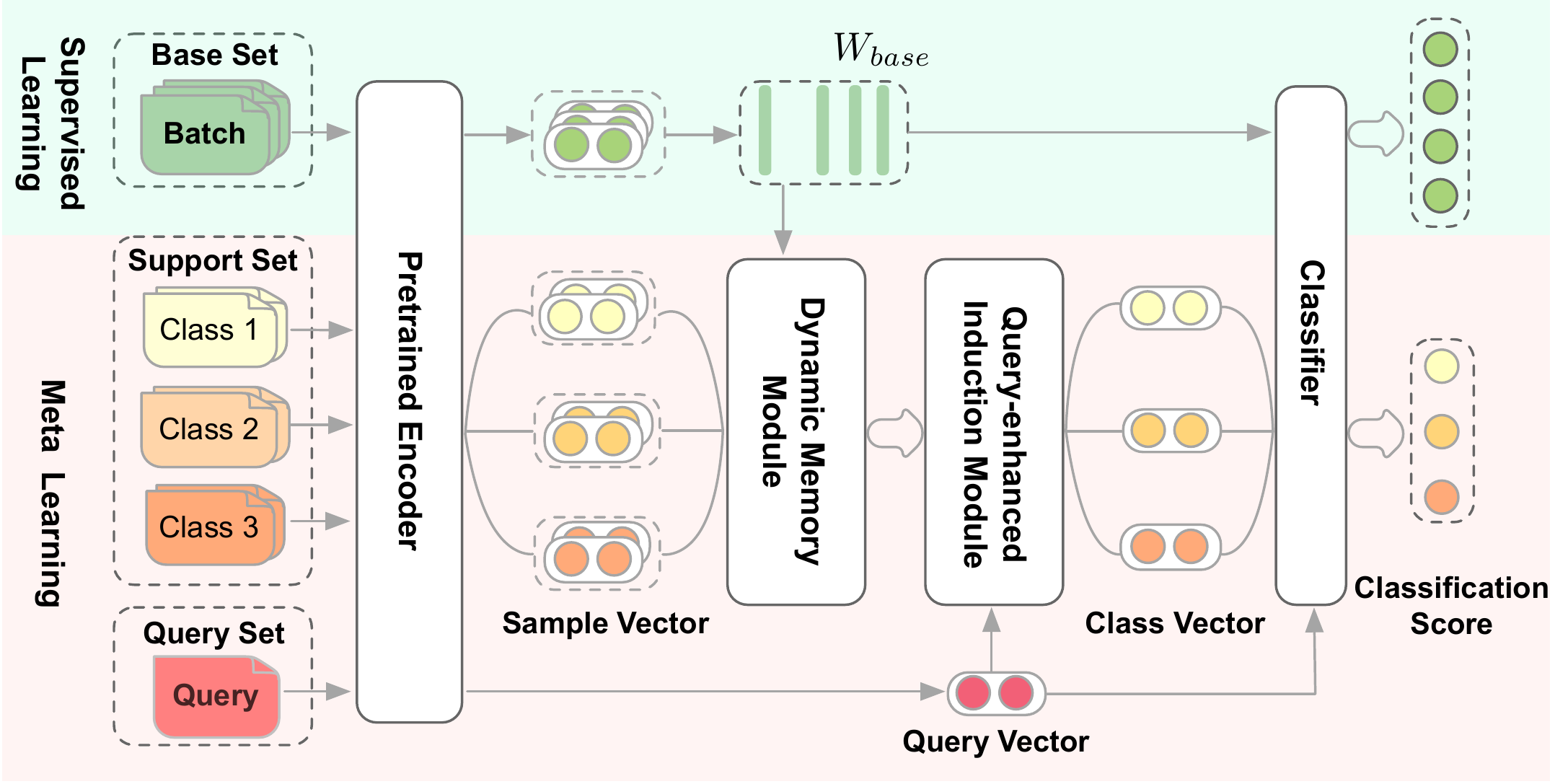}
\caption{An overview of Dynamic Memory Induction Network with a 3-way 2-shot example.} 
\centering
\label{model architechture}
\end{figure}

An overview of our Dynamic Memory Induction Networks (DMIN) is shown in Figure \ref{model architechture}, which is built on the two-stage few-shot framework~\citet{gidaris2018dynamic}. In the supervised learning stage (upper, green subfigure), a subset of classes in training data are selected as the base sets, consisting of $C_{base}$ number of base classes, which is used to finetune a pretrained sentence encoder and to train a classifier. 

In the meta-learning stage (bottom, orange subfigure), we construct an ``episode" to compute gradients and update our model in each training iteration. For a $C$-way $K$-shot problem, a training episode is formed by randomly selecting $C$ classes from the training set and choosing $K$ examples within each selected class to act as the support set $S = \cup_{c=1}^{C}\{x_{c,s},y_{c,s}\}_{s=1}^K$. A subset of the remaining examples serve as the query set $Q = \{x_{q},y_{q}\}_{q=1}^L$. 
Training on such episodes is conducted by feeding the support set $S$ to the model and updating its parameters to minimize the loss in the query set $Q$. 

\subsection{Pre-trained Encoder}
We expect that developing few-shot text classifier should benefit from the recent advance on pretrained models \citep{peters-etal-2018-deep,devlin2019bert,radford2018improving}. Unlike recent work~\citep{geng-etal-2019-induction}, we employ BERT-base \citep{devlin2019bert} for sentence encoding , which has been used in recent few-shot learning models \citep{bao2019few,soares2019matching}. The model architecture of BERT \citep{devlin2019bert} is a multi-layer bidirectional Transformer encoder based on the original Transformer model \citep{vaswani2017attention}. A special classification embedding ([CLS]) is inserted as the first token and a special token ([SEP]) is added as the final token. We use the $d$-dimensional hidden vector output from the $\left[CLS\right]$ as the representation $e$ of a given text $x$: $e=E(x|\theta)$.
The pre-trained BERT model provides a powerful context-dependent sentence representation and can be used for various target tasks, and it is suitable for the few-shot text classification task \citep{bao2019few,soares2019matching}. 

We finetune the pre-trained BERT encoder in the supervised learning stage. For each input document $x$, the encoder $E(x|\theta)$ (with parameter $\theta$) will output a vector $e$ of $d$ dimension. $W_{base}$ is a matrix that maintains a class-level vector for each base class, serving as a base memory for meta-learning. Both $E(x|\theta)$ and $W_{base}$ will be further tuned in the meta training procedure. We will show in our experiments that replacing previous models with pre-trained encoder outperforms the corresponding state-of-the-art models, and the proposed DMIN can further improve over that. 

\subsection{Dynamic Memory Module}
At the meta-learning stage, to induce class-level representations from given support sets, we develop a dynamic memory module (DMM) based on knowledge learned from the supervised learning stage through the memory matrix $W_{base}$. Unlike static memory~\citep{gidaris2018dynamic}, DMM utilizes dynamic routing \citep{sabour2017dynamic} to render more flexibility to the memory learned from base classes to better adapt support sets. The routing component can automatically adjust the coupling coefficients during and after training, which inherently suits for the need of few-shot learning. 


Specifically, the instances in the support sets are first encoded by the BERT into sample vectors $\{{e}_{c,s}\}_{s=1}^K$ and then fed to the following dynamic memory routing process.

\paragraph{Dynamic Memory Routing Process} The algorithm of the dynamic memory routing process, denoted as DMR, is presented in Algorighm~\ref{memory_routing_algorithm}. 

Given a memory matrix $M$ (here $W_{base}$) and sample vector $q \in {R^{d}}$, the algorithm aims to adapt the sample vector based on memory $M$ learned in the supervised learning stage.

\begin{equation}
    q^{\prime} = DMR(M, q).
\label{dynamic_memory_eq}
\end{equation}

First, for each entry ${m}_{i} \in M$, the standard matrix-transformation and squash operations in dynamic routing \citep{sabour2017dynamic} are applied on the inputs:
\begin{eqnarray}
{\hat{m}_{ij}} &=& squash({W_j}{{m}_{i}} + b_j),\label{squash_transformation}\\
{\hat{q}_{j}} &=& squash({W_j}{{q}} + b_j), \label{squash_transformation2}
\end{eqnarray}
where the transformation weights $W_j$ and bias $b_j$ are shared across the inputs to fit the few-shot learning scenario. 


We then calculate the Pearson Correlation Coefficients (PCCs) \citep{hunt1986percent, yang2019enhancing} between $\hat{m}_i$ and $\hat{q}_j$.
\begin{eqnarray}
p_{ij} &=& tanh(PCCs(\hat{m}_{ij}, \hat{q}_j)),\\
PCCs &=& \frac{Cov(x_1, x_2)}{\sigma_{x_1}\sigma_{x_2}}.
\end{eqnarray}
where the general formula of PCCs is given above for vectors $x_1$ and $x_2$. Since PCCs values are in the range of [-1, 1], they can be used to encourage or penalize the routing parameters.

\begin{algorithm}[t]
\algsetup{linenosize=\normalsize} \normalsize
\caption{Dynamic Memory Routing Process}
\begin{algorithmic}[1]
\REQUIRE $r$, $q$ and memory $M=\{m_1, m_2,...,m_n\}$
\ENSURE $\bm{v}={v_1, v_2,...,v_l}$, $q^{\prime}$
\FOR{all $m_i, v_j$}
\STATE $\hat{m}_{ij} = squash(W_{j} m_{i} + b_j)$
\STATE $\hat{q}_j = sqush(W_j q + b_j)$
\STATE $\alpha_{ij} = 0$
\STATE $p_{ij} = tanh(PCCs(\hat{m}_{ij}, \hat{q}_j))$
\ENDFOR
\FOR{ $r$ iterations }
\STATE ${d_i} = softmax \left( \alpha_i \right)$
\STATE $\hat{v}_j = \sum_{i=1}^n (d_{ij} + p_{ij}) \hat{m}_{ij}$
\STATE $v_j = squash(\hat{v}_j)$
\STATE for all $i,j$:  $\alpha_{ij} = \alpha_{i,j} + p_{ij} \hat{m}_{ij} v_j$
\STATE for all $j$:  $\hat{q}_j = \frac{\hat{q}_j + v_j}{2}$
\STATE for all $i, j$:  $p_{ij} = tanh(PCCs(\hat{m}_{ij}, \hat{q}_{j}))$
\ENDFOR
\STATE $q^{\prime} = concat[\bm{v}]$
\STATE \textbf{Return}  $q^{\prime}$
\end{algorithmic}
\label{memory_routing_algorithm}
\end{algorithm}

The routing iteration process can now adjust coupling coefficients, denoted as $d_i$, with regard to the input capsules $m_i$, $q$ and 
higher level capsules $v_j$. 
\begin{eqnarray}
{d_i} &=& softmax \left( \alpha_i \right ), \label{d:eq}\\
\alpha_{ij} &=& \alpha_{ij} + p_{ij} \hat{m}_i v_j. \label{alpha:eq}
\end{eqnarray}

Since our goal is to develop dynamic routing mechanism over memory for few-shot learning, we add the PCCs with the routing agreements in every routing iteration as shown in Eq. \ref{pcc_agree}.
\begin{eqnarray}
\hat{v}_j &=& \sum_{i=1}^n (d_{ij}+p_{ij})m_{ij}, \label{pcc_agree}\\
    v_j &=& squash(\hat{v}_j).
\end{eqnarray}
We update the coupling coefficients $\alpha_{ij}$ and $p_{ij}$ with Eq.~\ref{d:eq} and Eq.~\ref{alpha:eq}, and finally output the adapted vector $q^{\prime}$ as in Algorithm~\ref{memory_routing_algorithm}. 

The Dynamic Memory Module (DMM) aims to use DMR to adapt sample vectors $e_{c,s}$, guided by the memory $W_{base}$. That is, the resulting adapted sample vector is computed with  $e_{c,s}^{\prime} = DMR(W_{base}, e_{c,s})$.

\subsection{Query-enhanced Induction Module}
After the sample vectors $\left\{e_{c,s}^{\prime}\right\}_{s=1,...,K}$ are adapted and query vectors $\{e_q\}_{q=1}^L$ are encoded by the pretrained encoder, we now incorporate queries to build a Query-guided Induction Module (QIM).
The aim is to identify, among (adapted) sample vectors of support sets, the vectors that are more relevant to the query, in order to construct class-level vectors to better classify the query. Since dynamic routing can automatically adjusts the coupling coefficients to help enhance related (e.g., similar) queries and sample vectors, and penalizes unrelated ones, QIM reuses the DMR process by treating adapted sample vectors as memory of background knowledge about novel classes, and induces class-level representation from the adapted sample vectors that are more relevant/similar to the query under concern.
\begin{equation}
    e_c = DMR(\left\{e_{c,s}^{\prime}\right\}_{s=1,...,K}, e_q).
\end{equation}

\subsection{Similarity Classifier}
In the final classification stage, we then feed the novel class vector $e_c$ and query vector $e_q$ to the classifier discussed above in the supervised training stage and get the classification score. 
The standard setting for neural network classifiers is, after having extracted the feature vector $e \in R^d$, to estimate the classification probability vector $p$ by first computing the raw classification score $s_k$ of each category $k\in[1, K^*]$ using the dot-product operator $s_k=e^Tw_k^*$, and then applying softmax operator across all the $K^*$ classification scores. However, this type of classifiers do not fit few-shot learning due to completely novel categories. In this work, we compute the raw classification scores using a cosine similarity operator:
\begin{equation}
    s_k = \tau \cdot cos(e, w_k^*)=\tau \cdot \overline{e}^T \overline{w}_k^*,
    \label{cosine_base}
\end{equation}
where $\overline{e}=\frac{e}{\| e \| }$ and $\overline{w}_k^*=\frac{w_k^*}{\| w_k^* \| }$ are $l_2-$normalized vectors, and $\tau$ is a learnable scalar value.
After the base classifier is trained, all the feature vectors that belong to the same class must be very closely matched with the single classification weight vector of that class. So the base classification weights $W_{base} = \{ w_b \}_{b=1}^{C_{base}}$ trained in the $1$st stage can be seen as the base classes' feature vectors. 

In the few-shot classification scenario, we feed the query vector $e_q$ and novel class vector $e_c$ to the classifier and get the classification scores in a unified manner.
\begin{equation}
    s_{q,c} = \tau \cdot cos(e_q, e_c)=\tau \cdot \overline{e}_q^T \overline{e}_c.
    \label{cosine_few}
\end{equation}

\subsection{Objective Function}
In the supervised learning stage, the training objective is to minimize the cross-entropy loss on $C_{base}$ number of base classes given an input text $x$ and its label $\bm{y}$:
\begin{equation}
    L_1(x, \bm{y}, \bm{\hat{y}}) = - \sum\limits_{k=1}^{C_{base}} y_klog(\hat{y}_k),
\end{equation}
where $\bm{y}$ is one-hot representation of the ground truth label, and $\bm{\hat{y}}$ is the predicted probabilities of base classes with $\hat{y}_k=softmax(s_k)$.

In the meta-training stage, for each meta episode, given the support set $S$ and query set $Q =\{x_{q},y_{q}\}_{q=1}^L$, the training objective is to minimize the cross-entropy loss on $C$ novel classes.
\begin{equation}
    L_2(S, Q) = -\frac{1}{C} \sum\limits_{c=1}^{C} \frac{1}{L} \sum\limits_{q=1}^{L} y_{q}log(\hat{y}_{q}),
\end{equation}
where $\hat{y}_{q}=softmax(s_{q})$ is the predicted probabilities of $C$ novel classes in this meta episode, with $s_q=\{s_{q,c}\}_{c=1}^C$ from Equation \ref{cosine_few}. We feed the support set $S$ to the model and update its parameters to minimize the loss in the query set $Q$ in each meta episode. 

\section{Experiments}
\subsection{Dataset and Evaluation Metrics}
We evaluate our model on the miniRCV1~\citep{jiang2018attentive} and ODIC dataset \citep{geng-etal-2019-induction}. Following previous work~\citep{snell2017prototypical,geng-etal-2019-induction}, we use few-shot classification accuracy as the evaluation metric. We average over $100$ and $300$ randomly generated meta-episodes from the testing set in miniRCV1 and ODIC, respectively. We sample 10 test texts per class in each episode for evaluation in both the $1$-shot and $5$-shot scenarios. 
\subsection{Implementation Details}
We use Google pre-trained BERT-Base model as our text encoder, and fine-tune the model in the training procedure. The number of base classes $C_{base}$ on ODIC and miniRCV1 is set to be 100 and 20, respectively. The number of DMR interaction is 3. We build episode-based meta-training models with $C = [5, 10]$ and $K = [1, 5]$ for comparison. In addition to using K sample texts as the support set, the query set has 10 query texts for each of the C sampled classes in every training episode. For example, there are $10 \times 5 + 5 \times 5 = 75$ texts in one training episode for a 5-way 5-shot experiment.

\subsection{Results}
We compare DMIN with various baselines and state-of-the-art models: BERT \citep{devlin2019bert} finetune, ATAML \citep{jiang2018attentive}, Rel. Net \citep{sung2018learning}, Ind. Net \citep{geng-etal-2019-induction}, HATT \citep{gao2019hybrid}, and LwoF \citep{gidaris2018dynamic}. Note that we re-implement them with the BERT sentence encoder for direct comparison.

\begin{table}[t]  
\centering
\scalebox{0.7} {
\begin{tabular}{ccccc}  
\toprule
\multirow{2}*{\textbf{Model}} & \multicolumn{2}{c}{\textbf{5-way Acc.}} & \multicolumn{2}{c}{\textbf{10-way Acc.}}\\
~ & 1-shot & 5-shot & 1-shot & 5-shot \\
\midrule
BERT &  30.79$\pm$0.68 &  63.31$\pm$0.73 &  23.48$\pm0.53$ &  61.18$\pm$0.82 \\
ATAML & 54.05$\pm$0.14 & 72.79$\pm$0.27 & 39.48$\pm$0.23 & 61.74$\pm$0.36 \\
Rel. Net & 59.19$\pm$0.12 & 78.35$\pm$0.27 & 44.69$\pm$0.19 & 67.49$\pm$0.23 \\
Ind. Net & 60.97$\pm$0.16 & 80.91$\pm$0.19 & 46.15$\pm$0.26 & 69.42$\pm$0.34 \\
HATT & 60.40$\pm$0.17 & 79.46$\pm$0.32 & 47.09$\pm$0.28 & 68.58$\pm$0.37 \\
LwoF & 63.35$\pm$0.26 & 78.83$\pm$0.38 & 48.61$\pm$0.21 & 69.57$\pm$0.35 \\
\hline
DMIN & \textbf{65.72}$\pm$0.28 & \textbf{82.39}$\pm$0.24 & \textbf{49.54}$\pm$0.31 & \textbf{72.52}$\pm$0.25 \\
\bottomrule
\end{tabular}  
}
\caption{Comparison of accuracy (\%) on miniRCV1 with standard deviations.}
\label{RCV1 Results}
\end{table}

\begin{table}[t]
\centering
\scalebox{0.7}{
\begin{tabular}{ccccc}  
\toprule
\multirow{2}*{\textbf{Model}} & \multicolumn{2}{c}{\textbf{5-way Acc.}} & \multicolumn{2}{c}{\textbf{10-way Acc.}}\\
~ & 1-shot & 5-shot & 1-shot & 5-shot \\
\midrule
BERT &  38.06$\pm$0.27 &  64.24$\pm$0.36 &  29.24$\pm$0.19 &  64.53$\pm$0.35 \\
ATAML & 79.60$\pm$0.42 & 88.53$\pm$0.57 & 63.52$\pm$0.34 & 77.36$\pm$0.57\\
Rel. Net & 79.41$\pm$0.42 & 87.93$\pm$0.31 & 64.36$\pm$0.58 & 78.62$\pm$0.54\\
Ind. Net & 81.28$\pm$0.26 & 89.67$\pm$0.28 & 64.53$\pm$0.38 & 80.48$\pm$0.25\\
HATT & 81.57$\pm$0.47 & 89.27$\pm$0.58 & 65.75$\pm$0.61 & 81.53$\pm$0.56\\
LwoF & 79.52$\pm$0.29 & 87.34$\pm$0.34 & 65.04$\pm$0.43 & 80.69$\pm$0.37\\
\hline
DMIN & \textbf{83.46}$\pm$0.36 & \textbf{91.75}$\pm$0.23 & \textbf{67.31}$\pm$0.25 & \textbf{82.84}$\pm$0.38\\
\bottomrule
\end{tabular}
}
\caption{Comparison of accuracy(\%) on ODIC with standard deviations.}
\label{ODIC results}
\end{table}

\paragraph{Overall Performance}
The accuracy and standard deviations of the models are shown in Table~\ref{RCV1 Results} and ~\ref{ODIC results}. We can see that DMIN consistently outperform all existing models and achieve new state-of-the-art results on both datasets. The differences between DMIN and all the other models are statistically significant under the one-tailed paired t-test at the 95\% significance level. 

Note that LwoF builds a two-stage training procedure with a memory module learnt from the supervised learning and used in the meta-learning stage, but the memory mechanism is static after training, while DMIN uses dynamic memory routing to automatically adjust the coupling coefficients after training to generalize to novel classes, and outperform LwoF significantly. 
Note also that the performance of some of the baseline models (Rel. Net and Ind. Net) reported in Table~\ref{RCV1 Results} and ~\ref{ODIC results} is higher than that in~\citet{geng-etal-2019-induction} since we used BERT to replace BiLSTM-based encoders. The BERT encoder improves the baseline models by a powerful context meaning representation ability, and our model can further outperform these models with a dynamic memory routing method. Even with these stronger baselines, the proposed DMIN consistently outperforms them on both dataset.


\paragraph{Ablation Study}
We analyze the effect of different components of DMIN on ODIC in Table~\ref{ablation}. Specifically, we remove DMM and QIM, and vary the number of DMR iterations. We see that the best performance is achieved with $3$ iterations. The results show the effectiveness of both the dynamic memory module and the induction module with query information.



\begin{table}[t]  
\centering
\small
\scalebox{0.9}{
\begin{tabular}{cccc}  
\toprule
\textbf{Model} & \textbf{Iteration}& \textbf{1 Shot} & \textbf{5 Shot}\\ 
\midrule
w/o DMM & 3 & 81.79 & 90.19\\
w/o QIM & 3 & 82.37 & 90.57\\
DMIN & 1 & 82.70 & 90.92\\
DMIN & 2 & 82.95 & 91.18\\
DMIN & 3 & \textbf{83.46} & \textbf{91.75} \\
\bottomrule
\end{tabular}  
}
\caption{Ablation study of accuracy (\%) on ODIC in a 5-way setup.}
\label{ablation}
\end{table}  

\begin{figure}[t]
\centering
\includegraphics[width=7.5cm]{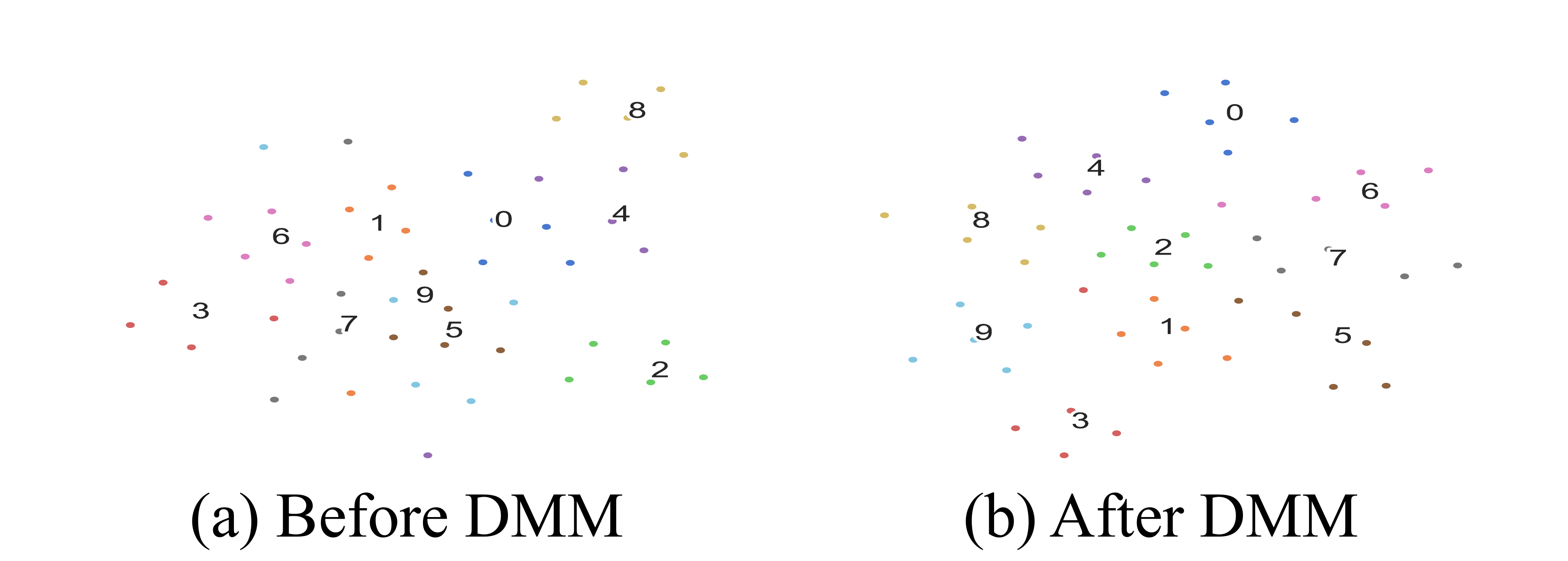}
\caption{Effect of the Dynamic Memory Module in a 10-way 5-shot setup.}
\label{fig:sandian}
\centering
\end{figure}

\subsection{Further Analysis}
Figure~\ref{fig:sandian} is the t-SNE visualization \citep{maaten2008visualizing} for support sample vectors before and after DMM under a $10$-way $5$-shot setup on ODIC. We randomly select a support set with 50 texts (10 texts per class) from the ODIC testing set, and obtain the sample vectors before and after DMM, i.e., ${\left\{ {{{e}_{c,s}}} \right\}_{c = 1,...5,s = 1...10}}$ and ${\left\{ {{e}_{c,s}^{\prime}} \right\}_{c = 1,...5,s = 1...10}}$. We can see that the support vectors produced by the DMM are better separated, demonstrating the effectiveness of DMM in leveraging the supervised learning experience to encode semantic relationships between lower level instance features and higher level class features for few-shot text classification.
\section{Conclusion}
We propose Dynamic Memory Induction Networks (DMIN) for few-shot text classification, which builds on external working memory with dynamic routing, leveraging the latter to track previous learning experience and the former to adapt and generalize better to support sets and hence to unseen classes. The model achieves new state-of-the-art results on the miniRCV1 and ODIC datasets. Since dynamic memory can be a learning mechanism more general than what we have used here for few-shot learning, we will investigate this type of models in other learning problems.

\section*{Acknowledgments}
The authors would like to thank the organizers of ACL-2020 and the reviewers for their helpful suggestions. The research of the last author is supported by the Natural Sciences and Engineering Research Council of Canada (NSERC).

\bibliography{acl2020}

\begin{thebibliography}{45}
\expandafter\ifx\csname natexlab\endcsname\relax\def\natexlab#1{#1}\fi

\bibitem[{Allen et~al.(2019)Allen, Shelhamer, Shin, and
  Tenenbaum}]{pmlr-v97-allen19b}
Kelsey Allen, Evan Shelhamer, Hanul Shin, and Joshua Tenenbaum. 2019.
\newblock \href {http://proceedings.mlr.press/v97/allen19b.html} {Infinite
  mixture prototypes for few-shot learning}.
\newblock In \emph{Proceedings of the 36th International Conference on Machine
  Learning}, volume~97 of \emph{Proceedings of Machine Learning Research},
  pages 232--241, Long Beach, California, USA. PMLR.

\bibitem[{Ba et~al.(2016)Ba, Hinton, Mnih, Leibo, and Ionescu}]{ba2016using}
Jimmy Ba, Geoffrey~E Hinton, Volodymyr Mnih, Joel~Z Leibo, and Catalin Ionescu.
  2016.
\newblock Using fast weights to attend to the recent past.
\newblock In \emph{Advances in Neural Information Processing Systems}, pages
  4331--4339.

\bibitem[{Bao et~al.(2019)Bao, Wu, Chang, and Barzilay}]{bao2019few}
Yujia Bao, Menghua Wu, Shiyu Chang, and Regina Barzilay. 2019.
\newblock Few-shot text classification with distributional signatures.
\newblock \emph{arXiv preprint arXiv:1908.06039}.

\bibitem[{Cai et~al.(2018)Cai, Pan, Yao, Yan, and Mei}]{cai2018memory}
Qi~Cai, Yingwei Pan, Ting Yao, Chenggang Yan, and Tao Mei. 2018.
\newblock Memory matching networks for one-shot image recognition.
\newblock In \emph{Proceedings of the IEEE Conference on Computer Vision and
  Pattern Recognition}, pages 4080--4088.

\bibitem[{Chen et~al.(2019)Chen, Zhang, Zhang, Chen, and
  Chen}]{chen-etal-2019-meta}
Mingyang Chen, Wen Zhang, Wei Zhang, Qiang Chen, and Huajun Chen. 2019.
\newblock \href {https://doi.org/10.18653/v1/D19-1431} {Meta relational
  learning for few-shot link prediction in knowledge graphs}.
\newblock In \emph{Proceedings of the 2019 Conference on Empirical Methods in
  Natural Language Processing and the 9th International Joint Conference on
  Natural Language Processing (EMNLP-IJCNLP)}, pages 4208--4217, Hong Kong,
  China. Association for Computational Linguistics.

\bibitem[{Das et~al.(2017)Das, Zaheer, Reddy, and
  McCallum}]{das-EtAl:2017:Short}
Rajarshi Das, Manzil Zaheer, Siva Reddy, and Andrew McCallum. 2017.
\newblock \href {http://aclweb.org/anthology/P17-2057} {Question answering on
  knowledge bases and text using universal schema and memory networks}.
\newblock In \emph{Proceedings of the 55th Annual Meeting of the Association
  for Computational Linguistics (Volume 2: Short Papers)}, pages 358--365,
  Vancouver, Canada. Association for Computational Linguistics.

\bibitem[{Devlin et~al.(2019)Devlin, Chang, Lee, and
  Toutanova}]{devlin2019bert}
Jacob Devlin, Ming-Wei Chang, Kenton Lee, and Kristina Toutanova. 2019.
\newblock Bert: Pre-training of deep bidirectional transformers for language
  understanding.
\newblock In \emph{Proceedings of the 2019 Conference of the North American
  Chapter of the Association for Computational Linguistics: Human Language
  Technologies, Volume 1 (Long and Short Papers)}, pages 4171--4186.

\bibitem[{Dou et~al.(2019)Dou, Yu, and Anastasopoulos}]{dou2019investigating}
Zi-Yi Dou, Keyi Yu, and Antonios Anastasopoulos. 2019.
\newblock Investigating meta-learning algorithms for low-resource natural
  language understanding tasks.
\newblock In \emph{Proceedings of the 2019 Conference on Empirical Methods in
  Natural Language Processing and the 9th International Joint Conference on
  Natural Language Processing (EMNLP-IJCNLP)}, pages 1192--1197.

\bibitem[{Fe-Fei et~al.(2003)}]{fe2003bayesian}
Li~Fe-Fei et~al. 2003.
\newblock A bayesian approach to unsupervised one-shot learning of object
  categories.
\newblock In \emph{Computer Vision, 2003. Proceedings. Ninth IEEE International
  Conference on}, pages 1134--1141. IEEE.

\bibitem[{Fei-Fei et~al.(2006)Fei-Fei, Fergus, and Perona}]{fei2006one}
Li~Fei-Fei, Rob Fergus, and Pietro Perona. 2006.
\newblock One-shot learning of object categories.
\newblock \emph{IEEE transactions on pattern analysis and machine
  intelligence}, 28(4):594--611.

\bibitem[{Finn et~al.(2017)Finn, Abbeel, and Levine}]{finn2017model}
Chelsea Finn, Pieter Abbeel, and Sergey Levine. 2017.
\newblock Model-agnostic meta-learning for fast adaptation of deep networks.
\newblock In \emph{Proceedings of the 34th International Conference on Machine
  Learning-Volume 70}, pages 1126--1135. JMLR. org.

\bibitem[{Gao et~al.(2019)Gao, Han, Liu, and Sun}]{gao2019hybrid}
Tianyu Gao, Xu~Han, Zhiyuan Liu, and Maosong Sun. 2019.
\newblock Hybrid attention-based prototypical networks for noisy few-shot
  relation classification.
\newblock In \emph{Proceedings of the Thirty-Second AAAI Conference on
  Artificial Intelligence,(AAAI-19), New York, USA}.

\bibitem[{Geng et~al.(2019)Geng, Li, Li, Zhu, Jian, and
  Sun}]{geng-etal-2019-induction}
Ruiying Geng, Binhua Li, Yongbin Li, Xiaodan Zhu, Ping Jian, and Jian Sun.
  2019.
\newblock \href {https://doi.org/10.18653/v1/D19-1403} {Induction networks for
  few-shot text classification}.
\newblock In \emph{Proceedings of the 2019 Conference on Empirical Methods in
  Natural Language Processing and the 9th International Joint Conference on
  Natural Language Processing (EMNLP-IJCNLP)}, pages 3895--3904, Hong Kong,
  China. Association for Computational Linguistics.

\bibitem[{Gidaris and Komodakis(2018)}]{gidaris2018dynamic}
Spyros Gidaris and Nikos Komodakis. 2018.
\newblock Dynamic few-shot visual learning without forgetting.
\newblock In \emph{Proceedings of the IEEE Conference on Computer Vision and
  Pattern Recognition}, pages 4367--4375.

\bibitem[{Gu et~al.(2018)Gu, Wang, Chen, Li, and Cho}]{gu2018meta}
Jiatao Gu, Yong Wang, Yun Chen, Victor~OK Li, and Kyunghyun Cho. 2018.
\newblock Meta-learning for low-resource neural machine translation.
\newblock In \emph{Proceedings of the 2018 Conference on Empirical Methods in
  Natural Language Processing}, pages 3622--3631.

\bibitem[{Hu et~al.(2019)Hu, Chen, Chang, and Sun}]{hu-etal-2019-shot}
Ziniu Hu, Ting Chen, Kai-Wei Chang, and Yizhou Sun. 2019.
\newblock \href {https://doi.org/10.18653/v1/P19-1402} {Few-shot representation
  learning for out-of-vocabulary words}.
\newblock In \emph{Proceedings of the 57th Annual Meeting of the Association
  for Computational Linguistics}, pages 4102--4112, Florence, Italy.
  Association for Computational Linguistics.

\bibitem[{Hunt(1986)}]{hunt1986percent}
Ronald~J Hunt. 1986.
\newblock Percent agreement, pearson's correlation, and kappa as measures of
  inter-examiner reliability.
\newblock \emph{Journal of Dental Research}, 65(2):128--130.

\bibitem[{Jiang et~al.(2018)Jiang, Havaei, Chartrand, Chouaib, Vincent, Jesson,
  Chapados, and Matwin}]{jiang2018attentive}
Xiang Jiang, Mohammad Havaei, Gabriel Chartrand, Hassan Chouaib, Thomas
  Vincent, Andrew Jesson, Nicolas Chapados, and Stan Matwin. 2018.
\newblock Attentive task-agnostic meta-learning for few-shot text
  classification.

\bibitem[{Kaiser et~al.(2017)Kaiser, Nachum, Roy, and
  Bengio}]{kaiser2017learning}
{\L}ukasz Kaiser, Ofir Nachum, Aurko Roy, and Samy Bengio. 2017.
\newblock Learning to remember rare events.
\newblock \emph{arXiv preprint arXiv:1703.03129}.

\bibitem[{Maaten and Hinton(2008)}]{maaten2008visualizing}
Laurens van~der Maaten and Geoffrey Hinton. 2008.
\newblock Visualizing data using t-sne.
\newblock \emph{Journal of machine learning research}, 9(Nov):2579--2605.

\bibitem[{Madotto et~al.(2018)Madotto, Wu, and Fung}]{madotto2018mem2seq}
Andrea Madotto, Chien-Sheng Wu, and Pascale Fung. 2018.
\newblock Mem2seq: Effectively incorporating knowledge bases into end-to-end
  task-oriented dialog systems.
\newblock In \emph{Proceedings of the 56th Annual Meeting of the Association
  for Computational Linguistics (Volume 1: Long Papers)}, pages 1468--1478.

\bibitem[{Mishra et~al.(2017)Mishra, Rohaninejad, Chen, and
  Abbeel}]{mishra2017simple}
Nikhil Mishra, Mostafa Rohaninejad, Xi~Chen, and Pieter Abbeel. 2017.
\newblock A simple neural attentive meta-learner.
\newblock \emph{arXiv preprint arXiv:1707.03141}.

\bibitem[{Munkhdalai and Yu(2017)}]{munkhdalai2017meta}
Tsendsuren Munkhdalai and Hong Yu. 2017.
\newblock Meta networks.
\newblock \emph{arXiv preprint arXiv:1703.00837}.

\bibitem[{Obamuyide and Vlachos(2019)}]{obamuyide-vlachos-2019-model}
Abiola Obamuyide and Andreas Vlachos. 2019.
\newblock \href {https://doi.org/10.18653/v1/P19-1589} {Model-agnostic
  meta-learning for relation classification with limited supervision}.
\newblock In \emph{Proceedings of the 57th Annual Meeting of the Association
  for Computational Linguistics}, pages 5873--5879, Florence, Italy.
  Association for Computational Linguistics.

\bibitem[{Peters et~al.(2018)Peters, Neumann, Iyyer, Gardner, Clark, Lee, and
  Zettlemoyer}]{peters-etal-2018-deep}
Matthew Peters, Mark Neumann, Mohit Iyyer, Matt Gardner, Christopher Clark,
  Kenton Lee, and Luke Zettlemoyer. 2018.
\newblock \href {https://doi.org/10.18653/v1/N18-1202} {Deep contextualized
  word representations}.
\newblock In \emph{Proceedings of the 2018 Conference of the North {A}merican
  Chapter of the Association for Computational Linguistics: Human Language
  Technologies, Volume 1 (Long Papers)}, pages 2227--2237, New Orleans,
  Louisiana. Association for Computational Linguistics.

\bibitem[{Qi et~al.(2018)Qi, Brown, and Lowe}]{qi2018low}
Hang Qi, Matthew Brown, and David~G Lowe. 2018.
\newblock Low-shot learning with imprinted weights.
\newblock In \emph{Proceedings of the IEEE Conference on Computer Vision and
  Pattern Recognition}, pages 5822--5830.

\bibitem[{Radford et~al.()Radford, Narasimhan, Salimans, and
  Sutskever}]{radford2018improving}
Alec Radford, Karthik Narasimhan, Tim Salimans, and Ilya Sutskever.
\newblock Improving language understanding by generative pre-training.

\bibitem[{Ravi and Larochelle(2016)}]{ravi2016optimization}
Sachin Ravi and Hugo Larochelle. 2016.
\newblock Optimization as a model for few-shot learning.

\bibitem[{Rios and Kavuluru(2018)}]{rios2018few}
Anthony Rios and Ramakanth Kavuluru. 2018.
\newblock Few-shot and zero-shot multi-label learning for structured label
  spaces.
\newblock In \emph{Proceedings of the 2018 Conference on Empirical Methods in
  Natural Language Processing}, pages 3132--3142.

\bibitem[{Sabour et~al.(2017)Sabour, Frosst, and Hinton}]{sabour2017dynamic}
Sara Sabour, Nicholas Frosst, and Geoffrey~E Hinton. 2017.
\newblock Dynamic routing between capsules.
\newblock In \emph{Advances in Neural Information Processing Systems}, pages
  3856--3866.

\bibitem[{Salamon and Bello(2017)}]{salamon2017deep}
Justin Salamon and Juan~Pablo Bello. 2017.
\newblock Deep convolutional neural networks and data augmentation for
  environmental sound classification.
\newblock \emph{IEEE Signal Processing Letters}, 24(3):279--283.

\bibitem[{Santoro et~al.(2016)Santoro, Bartunov, Botvinick, Wierstra, and
  Lillicrap}]{santoro2016meta}
Adam Santoro, Sergey Bartunov, Matthew Botvinick, Daan Wierstra, and Timothy
  Lillicrap. 2016.
\newblock Meta-learning with memory-augmented neural networks.
\newblock In \emph{International conference on machine learning}, pages
  1842--1850.

\bibitem[{Snell et~al.(2017)Snell, Swersky, and Zemel}]{snell2017prototypical}
Jake Snell, Kevin Swersky, and Richard Zemel. 2017.
\newblock Prototypical networks for few-shot learning.
\newblock In \emph{Advances in Neural Information Processing Systems}, pages
  4077--4087.

\bibitem[{Soares et~al.(2019)Soares, FitzGerald, Ling, and
  Kwiatkowski}]{soares2019matching}
Livio~Baldini Soares, Nicholas FitzGerald, Jeffrey Ling, and Tom Kwiatkowski.
  2019.
\newblock Matching the blanks: Distributional similarity for relation learning.
\newblock \emph{arXiv preprint arXiv:1906.03158}.

\bibitem[{Sun et~al.(2019)Sun, Liu, Chua, and Schiele}]{sun2019meta}
Qianru Sun, Yaoyao Liu, Tat-Seng Chua, and Bernt Schiele. 2019.
\newblock Meta-transfer learning for few-shot learning.
\newblock In \emph{Proceedings of the IEEE Conference on Computer Vision and
  Pattern Recognition}, pages 403--412.

\bibitem[{Sung et~al.(2018)Sung, Yang, Zhang, Xiang, Torr, and
  Hospedales}]{sung2018learning}
Flood Sung, Yongxin Yang, Li~Zhang, Tao Xiang, Philip~HS Torr, and Timothy~M
  Hospedales. 2018.
\newblock Learning to compare: Relation network for few-shot learning.
\newblock In \emph{Proceedings of the IEEE Conference on Computer Vision and
  Pattern Recognition}, pages 1199--1208.

\bibitem[{Tang et~al.(2016)Tang, Qin, and Liu}]{tang2016aspect}
Duyu Tang, Bing Qin, and Ting Liu. 2016.
\newblock Aspect level sentiment classification with deep memory network.
\newblock In \emph{Proceedings of the 2016 Conference on Empirical Methods in
  Natural Language Processing}, pages 214--224.

\bibitem[{Vaswani et~al.(2017)Vaswani, Shazeer, Parmar, Uszkoreit, Jones,
  Gomez, Kaiser, and Polosukhin}]{vaswani2017attention}
Ashish Vaswani, Noam Shazeer, Niki Parmar, Jakob Uszkoreit, Llion Jones,
  Aidan~N Gomez, {\L}ukasz Kaiser, and Illia Polosukhin. 2017.
\newblock Attention is all you need.
\newblock In \emph{Advances in Neural Information Processing Systems}, pages
  5998--6008.

\bibitem[{Vinyals et~al.(2016)Vinyals, Blundell, Lillicrap, Wierstra
  et~al.}]{vinyals2016matching}
Oriol Vinyals, Charles Blundell, Tim Lillicrap, Daan Wierstra, et~al. 2016.
\newblock Matching networks for one shot learning.
\newblock In \emph{Advances in Neural Information Processing Systems}, pages
  3630--3638.

\bibitem[{Wu et~al.(2019)Wu, Xiong, and Wang}]{wu2019learning}
Jiawei Wu, Wenhan Xiong, and William~Yang Wang. 2019.
\newblock Learning to learn and predict: A meta-learning approach for
  multi-label classification.
\newblock In \emph{Proceedings of the 2019 Conference on Empirical Methods in
  Natural Language Processing and the 9th International Joint Conference on
  Natural Language Processing (EMNLP-IJCNLP)}, pages 4345--4355.

\bibitem[{Xu et~al.(2019)Xu, Liu, Shu, and Yu}]{xu2019open}
Hu~Xu, Bing Liu, Lei Shu, and P~Yu. 2019.
\newblock Open-world learning and application to product classification.
\newblock In \emph{The World Wide Web Conference}, pages 3413--3419. ACM.

\bibitem[{Yang et~al.(2019)Yang, Zhang, Meng, Gu, Feng, and
  Zhou}]{yang2019enhancing}
Zhengxin Yang, Jinchao Zhang, Fandong Meng, Shuhao Gu, Yang Feng, and Jie Zhou.
  2019.
\newblock Enhancing context modeling with a query-guided capsule network for
  document-level translation.
\newblock In \emph{Proceedings of the 2019 Conference on Empirical Methods in
  Natural Language Processing and the 9th International Joint Conference on
  Natural Language Processing (EMNLP-IJCNLP)}, pages 1527--1537.

\bibitem[{Ye and Ling(2019)}]{ye-ling-2019-multi}
Zhi-Xiu Ye and Zhen-Hua Ling. 2019.
\newblock \href {https://doi.org/10.18653/v1/P19-1277} {Multi-level matching
  and aggregation network for few-shot relation classification}.
\newblock In \emph{Proceedings of the 57th Annual Meeting of the Association
  for Computational Linguistics}, pages 2872--2881, Florence, Italy.
  Association for Computational Linguistics.

\bibitem[{Yu et~al.(2018)Yu, Guo, Yi, Chang, Potdar, Cheng, Tesauro, Wang, and
  Zhou}]{yu2018diverse}
Mo~Yu, Xiaoxiao Guo, Jinfeng Yi, Shiyu Chang, Saloni Potdar, Yu~Cheng, Gerald
  Tesauro, Haoyu Wang, and Bowen Zhou. 2018.
\newblock Diverse few-shot text classification with multiple metrics.
\newblock In \emph{Proceedings of the 2018 Conference of the North American
  Chapter of the Association for Computational Linguistics: Human Language
  Technologies, Volume 1 (Long Papers)}, pages 1206--1215.

\bibitem[{Zhang et~al.(2018)Zhang, Che, Ghahramani, Bengio, and
  Song}]{zhang2018metagan}
Ruixiang Zhang, Tong Che, Zoubin Ghahramani, Yoshua Bengio, and Yangqiu Song.
  2018.
\newblock Metagan: An adversarial approach to few-shot learning.
\newblock In \emph{Advances in Neural Information Processing Systems}, pages
  2365--2374.

\end{thebibliography}
\bibliographystyle{acl_natbib}

\end{document}